\documentclass[conference,a4paper]{IEEEtran}
\IEEEoverridecommandlockouts

\usepackage[hidelinks]{hyperref}
\usepackage[cmex10]{amsmath}%American Math Society(AMS) math formatting
\usepackage{amssymb,amsfonts}%AMS extra symbols and fonts
\usepackage{dblfloatfix}%fix double column figure ordering and placement

\usepackage[ruled,vlined]{algorithm2e}
\usepackage{graphicx}
\graphicspath{{Figures/PDF/}{Figures/PNG/}}

\usepackage{booktabs}
\usepackage{siunitx}
\usepackage[numbers,compress]{natbib}
\usepackage{texnames}
\usepackage{bm,bbm}
\usepackage{orcidlink}

\usepackage{xcolor}

\definecolor{highlightcolor}{RGB}{0,0,255} % 蓝色

\hypersetup{
    colorlinks=true,       % 激活链接颜色
    citecolor=highlightcolor, % 引用颜色
    linkcolor=highlightcolor, % 内部链接颜色
    urlcolor=highlightcolor   % 外部链接颜色
}

\begin{document}

\title{\uppercase{Superpixel-Based Building Damage Detection from Post-earthquake Imagery Using Deep Neural Networks}
% Post-earthquake
}

\author{	
\IEEEauthorblockN{Jun Wang$^{1,2}$}
	\IEEEauthorblockA{\textit{1.Department of Networks, China Mobile Communications Group Co.,Ltd}, Beijing, 100032, China.\\	\textit{2.Institute of Remote Sensing and Geographic Information System, Peking University}, Beijing, 100871, China.\\
		}
	% \and
	% \IEEEauthorblockN{Hui Zhang\orcidlink{0000-0002-5283-7350}}
	% \IEEEauthorblockA{\textit{Inner Mongolia University}\\
	% 	010021 Hohhot, China\\
	% 	hui.zhang@imu.edu.cn}
	% \and
	% \IEEEauthorblockN{Andrea Rey\orcidlink{0000-0002-9185-1382}}
	% \IEEEauthorblockA{\textit{Universidad Nacional de Hurlingham}\\
	% 	1688 República Argentina\\
	% 	andrea.rey@unahur.edu.ar}
}

\maketitle
\begin{abstract}
	Building damage detection after natural disasters like earthquakes is crucial for initiating effective emergency response actions. Remotely sensed very high spatial resolution (VHR) imagery can provide vital information due to their ability to map the affected buildings with high geometric precision. However, we suffer from suboptimal performances in detecting damaged buildings due to earthquakes.
    % little attention has been paid to exploiting rich features represented in VHR images using Deep Neural Networks (DNN).
    This paper presents a novel superpixel based approach incorporates Deep Neural Networks (DNN) with a modified segmentation method, for more precise building damage detection from VHR imagery. Firstly, a modified Fast Scanning and Adaptive Merging method is extended to create initial over-segmentation. Secondly, the segments are properly merged based on the Region Adjacent Graph (RAG).
    % , considered an improved semantic similarity criterion composed of Local Binary Patterns (LBP) texture, spectral, and shape features
    Thirdly, a pre-trained DNN using Stacked Denoising Auto-Encoders (SDAE-DNN) is presented, to exploit the rich semantic features for building damage detection. Experimental results on a WorldView-2 imagery from Nepal Earthquake of 2015 demonstrate the feasibility and effectiveness of our method, which could boost detection accuracy through learning more intrinsic and discriminative features, which outperforms other methods using alternative classifiers.
\end{abstract}


\section{Introduction}

Rapid damage assessment after the earthquake is crucial for initiating effective emergency response actions. Remote sensing can quickly provide regional information about the affected areas with high thematic precision \cite{brunner2010earthquake,qing2022operational}. However, accurate and reliable damage detection remains a daunting challenge due to the complicated background interference, shooting angle, shadows, and solar illumination conditions \cite{derksen2020geometry}. In reality, building damage detection from very high resolution (VHR) remotely sensed imagery is one of the most challenging topics in emergency response and disaster management \cite{ye2017building}.

Valuable pre-earthquake data are always not guaranteed shortly after an event. Consequently, building damage detection using only post-earthquake VHR images plays a significant role in rapid damage assessment during the ﬁrst response \cite{brunner2010earthquake,dong2013comprehensive}. More precisely, post-earthquake VHR image classification based on various image features is the most commonly adopted strategy \cite{dong2013comprehensive}. Along with image classification moving from pixel to object, Blaschke et al. \cite{blaschke2014geographic} have elucidated the shortcomings of the pure per-pixel classification approach, which leads to splitting meaningful objects into separated pixels without explicit semantic features. Whereas, in terms of the per-superpixel strategy, Geographic Object-Based Image Analysis (GEOBIA) \cite{blaschke2014geographic} provides a general architecture that is capable of segmenting images and extracting features over objects of interest, it avoids the intrinsic drawbacks of pixel-based methodology. In general, superpixel segmentation methods can be categorized into two types according to the features used: edge-based methods and region-based methods \cite{blaschke2014geographic}. Edge-based methods take boundary as the primary vision clue for partition image contents, without considering regional features \cite{meng2016globally}. On the other hand, region-based methods split or merge neighboring pixels according to their regional semantic homogeneity and heterogeneity. However, most region-based segmentations like Fractal Net Evolution Approach (FNEA) \cite{Baatz2000MultiresolutionS} embedded in eCognition suffered from inconsistent segmentation for utilizing merely spectral and shape features, but neglected the key visual cues like texture. Therefore, to address the lack of exploration on rich semantic features like texture, we need to incorporate textural semantics over the techniques mentioned above to obtain more reasonable segmentation.  

\begin{figure*}[!ht]
\centering
\includegraphics[width=0.92\linewidth]{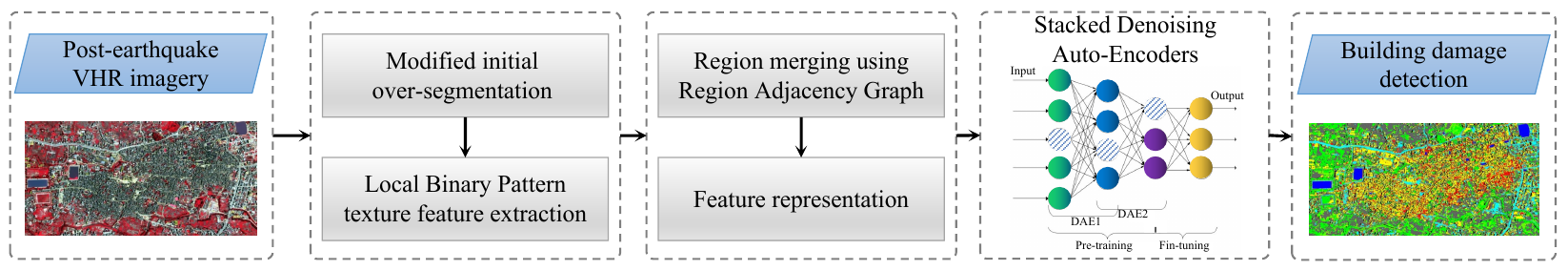}
\vspace{-0.5cm}
\caption{Flowchart of the proposed building damage detection method. } 
\label{fig:frame}
\end{figure*}

Based on the per-superpixel strategy, Liu et al. \cite{liu2015hyperspectral} used simple linear iterative clustering (SLIC) \cite{achanta2012slic} to generate superpixels for hyperspectral classification, but SLIC had notable drawbacks in short of shape matching. Moreover, it's worth noting that they stated in \cite{liu2015hyperspectral} that the segmentation was only treated as spatial constraints, but not involved in feature representation stage to exploit the rich regional semantic features. More importantly, besides the image segmentation results, the subsequent feature representation of semantic objects plays an utmost critical role in detection. To the best of our knowledge, the detection or classification tasks contain two critical steps: feature design by prior human knowledge and machine learning by classifiers. However, most of the existing building damage detection methods utilize conventional classifiers,  such as Support Vector Machine (SVM) \cite{khoshelham2013segment}, Random Forests (RF) \cite{thomas2013automated}, Extreme Learning Machine (ELM) \cite{alhichri2015hierarchical} and Multi-Layer Perceptron (MLP) \cite{dubois2014fast}, these methods always encounter the main weakness arising from insufficient feature learning \cite{khoshelham2013segment}, and suffer from tedious feature designing and selection. Recent research has shown that the performance of machine learning methods heavily depends on feature representation and learning \cite{hinton2006fast,berahmand2024autoencoders,gou2024hierarchical}. In this case, designing or finding good features is the critical problem of developing a proper classifier-based strategy. Alternatively, unlike conventional frameworks, DNN or deep learning architecture could manipulate the hand-craft feature design process into a machine self-learning manner \cite{bengio2013representation}. Consequently, feature learning and pattern discrimination can be achieved by DNN itself jointly.

\vspace{-0.5cm}

Given this, the main contributions of this paper are twofold: (1) Firstly, a texture-enhanced segmentation method is proposed to generate more meaningful superpixels. (2) Secondly, we proposed a SDAE-DNN framework to combine the benefit of superpixel segmentation and deep feature learning architecture for optimal damage building detection performance.

\section{BUILDING DAMAGE DETECTION WITH DNN}

Based on the approaches mentioned above, the intuition of our paper is to develop a superpixel-based method for building damage detection to address the issue of semantic feature exploration. A framework combining image segmentation and deep neural networks is proposed for building damage detection from post-earthquake VHR optical imagery. The flowchart of the proposed approach is summarized in Fig. \ref{fig:frame}, which is structured as follows: 

1) The VHR imagery is initially partitioned by modifying the Fast Scanning and Adaptive Merging (FSAM) algorithm \cite{ding2009efficient}, Meanwhile, the semantic features of each superpixel include Local Binary Pattern (LBP) texture \cite{ojala2002multiresolution} are calculated. 

2) Then, a region merging criterion integrating texture feature is proposed, more perceptually meaningful superpixels are generated based on Region Adjacent Graph (RAG). Then various image features of merged superpixels are extracted to represent the semantic information of regions.

3) On the basis of feature extraction on previously generated superpixels, a pre-trained DNN using Stacked Denoising AutoEncoders \cite{vincent2010stacked} (SDAE-DNN) was presented to classify and detect superpixels that indicate affected buildings, the results are evaluated and compared with state-of-the-art alternative methods in the 2015 Nepal earthquake.

\vspace{-0.5cm}

\subsection{Initial partition using modified FSAM}

Firstly, the VHR imagery was over-segmented by modifying the Fast Scanning and Adaptive Merging (FSAM) algorithm \cite{khoshelham2013segment}. FSAM was proved to be an efficient initial segmentation algorithm with good shape connectivity and shape matching. Nevertheless, FSAM was originally designed for only single-band images, which paid less attention to excavating the rich spectral information of multi-spectral VHR images. Therefore, spectral angle mapping (SAM) \cite{liu2019classification} was applied to customize for VHR multi-band imagery in our study, so as to make full use of the rich spectral features. The homogeneity between region $R_{obj1}$ and $R_{obj2}$ is calculated as:

\begin{equation}
\begin{aligned}
h_{SAM}\left(R_{obj1},R_{obj2}\right) = cos^{-1}\left( \frac{\theta_{obj1}\cdot\theta_{obj2}}{\left \| \theta_{obj1} \right \|_2 \left \| \theta_{obj2} \right \|_2}\right) \\ 
=\cos^{-1}{\left(\frac{\sum_{b=1}^{B}{\theta_{obj1}\cdot\theta_{obj2}}}{\left(\sum_{b=1}^{B}{\theta_{obj1}}^2\right)^{1/2}\left(\sum_{b=1}^{B}{\theta_{obj2}}^2\right)^{1/2}}\right)}
\end{aligned}
\end{equation}

where $\theta_{obj1}$ and $\theta_{obj2}$ represent the spectral vector of adjacent obj1 and obj2, respectively. $B$ represents the total band number of the input image.

\begin{figure}[!ht]
\centering
\includegraphics[width=0.9\linewidth]{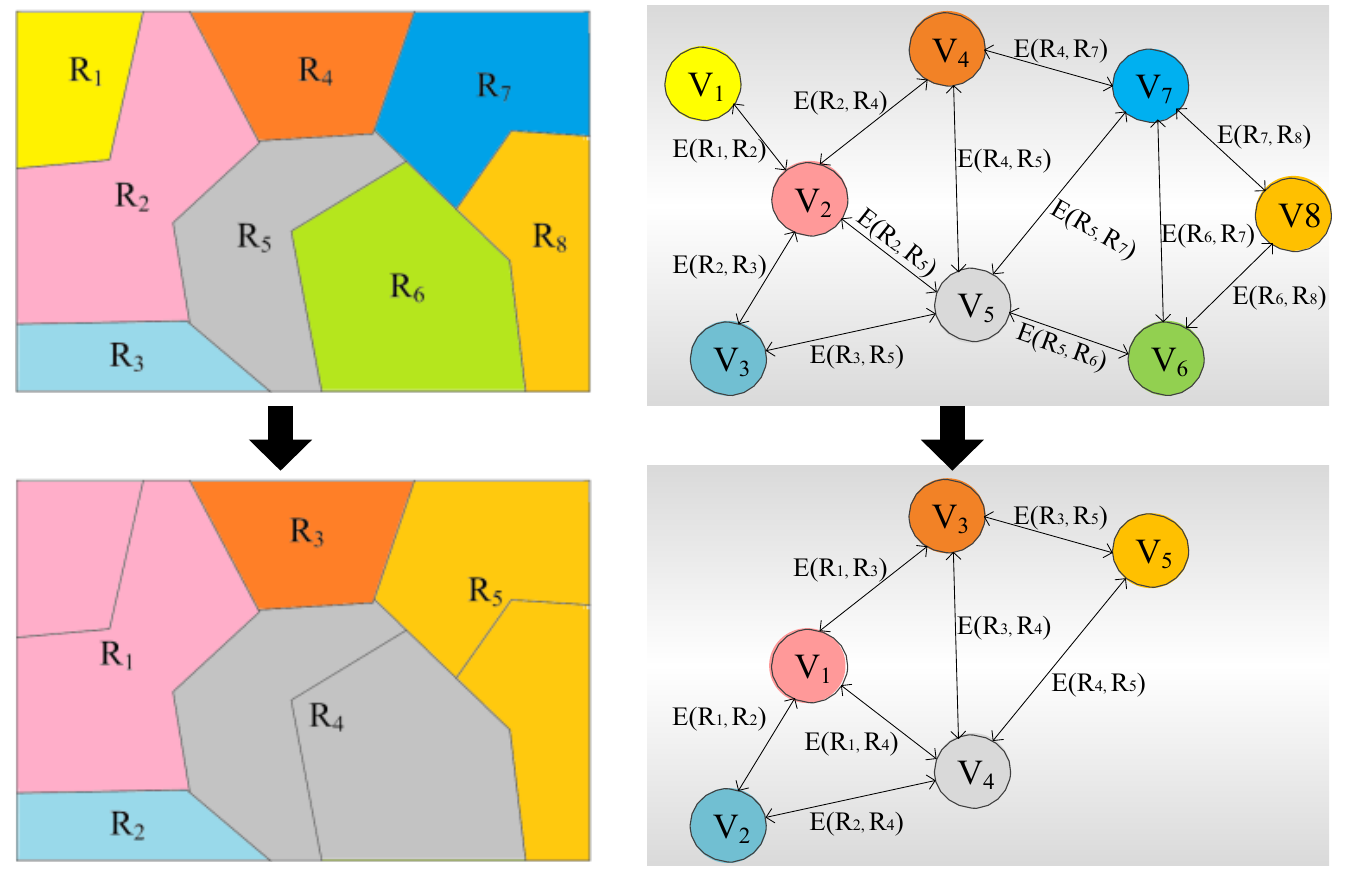}
\vspace{-0.5cm}
\caption{Region merging based on RAG. Left: Image region merge. Right: Corresponding Region Adjacent Graph (RAG). } 
\label{fig:RAG}
\end{figure}

\vspace{-0.4cm}

\subsection{Region Merging based on RAG}

In this section, an improved region merging criterion incorporating texture information was proposed to group the pixels into perceptually meaningful geospatial regions. 
Regions of the initial partition are represented as nodes of a corresponding RAG, as illustrated in Fig. \ref{fig:RAG}. The weights of the edges represent the heterogeneity between the adjacent pair of regions. A minimum heterogeneity criterion \cite{Baatz2000MultiresolutionS} was applied to RAG to find the optimal segmentation. During the merging, texture clustering can refine the real edges of the texture region. More precisely, the semantic heterogeneity consisted of spectral, textual and geometric information. It was defined as: 

\begin{equation}
\begin{aligned}
h=\omega_{spec}\ast h_{spec}+\omega_{texture}\ast h_{texture}+\omega_{shape}\ast h_{shape}
\end{aligned}
\end{equation}

where $h_\text{spec}$, $h_\text{texture}$ and $h_\text{shape}$ represented the heterogeneity derived from spectral, texture and shape feature respectively. $W_\text{spec}$, $W_\text{texture}$ and $W_\text{shape}$ indicated the weights of spectral, texture and shape feature respectively, and $W_\text{spec}$, $W_\text{texture}$, $W_\text{shape}$ $\in [0,1]$, $W_\text{spec}$ + $W_\text{texture}$ + $W_\text{shape}$ = 1. Spectral/ texture/shape weights were set to 0.7/0.2/0.1, because generally spectral information plays a more primary role in segmentation. The shape heterogeneity $h_\text{shape}$ describes the difference of compact degree $h_\text{compact}$ and smooth degree $h_\text{smooth}$ before and after adjacent region merging, the details of $h_\text{compact}$ and $h_\text{shape}$ are clearly described in \cite{Baatz2000MultiresolutionS}. Smoothness/compactness weights for shape feature were set to 0.5/0.5 because we did not want to favor either compact or non-compact superpixels. As the experimental image has 8 bands, the spectral heterogeneity was also calculated using SAM defined in Equation (1).

Besides spectral and shape features, a theoretically simple but effective texture descriptor named local binary pattern (LBP) has been proposed and widely used in computer vision \cite{ojala2002multiresolution}. Therefore, in our study, the joint distribution of LBP was used to characterize textural features over adjacent regions generated by the previous modified FSAM algorithm.

The initial modified FSAM segmentation and minimum heterogeneity region merging results are illustrated in Fig. \ref{fig:RAG}. The following elementary features were calculated for each superpixel: 1) Spectral feature (mean values and variance for each band); 2) Shape feature (area, shape index, length/width, rectangular ﬁt, roundness and density), which are calculated to distinguish between buildings and roads; 3) Texture feature (The near-infrared reflectance (NIR) band was used for grey-level co-occurrence matrix (GLCM) contrast, correlation, and entropy (all directions), considering that the NIR band is one of the most informative bands in VHR images for urban land cover classification). Besides, 4) mean normalized difference vegetation index (NDVI) of each segment was also calculated to handle vegetations and non-vegetations.

\vspace{-0.4cm}

\begin{figure}[!h]
\centering
\includegraphics[width=0.85\linewidth]{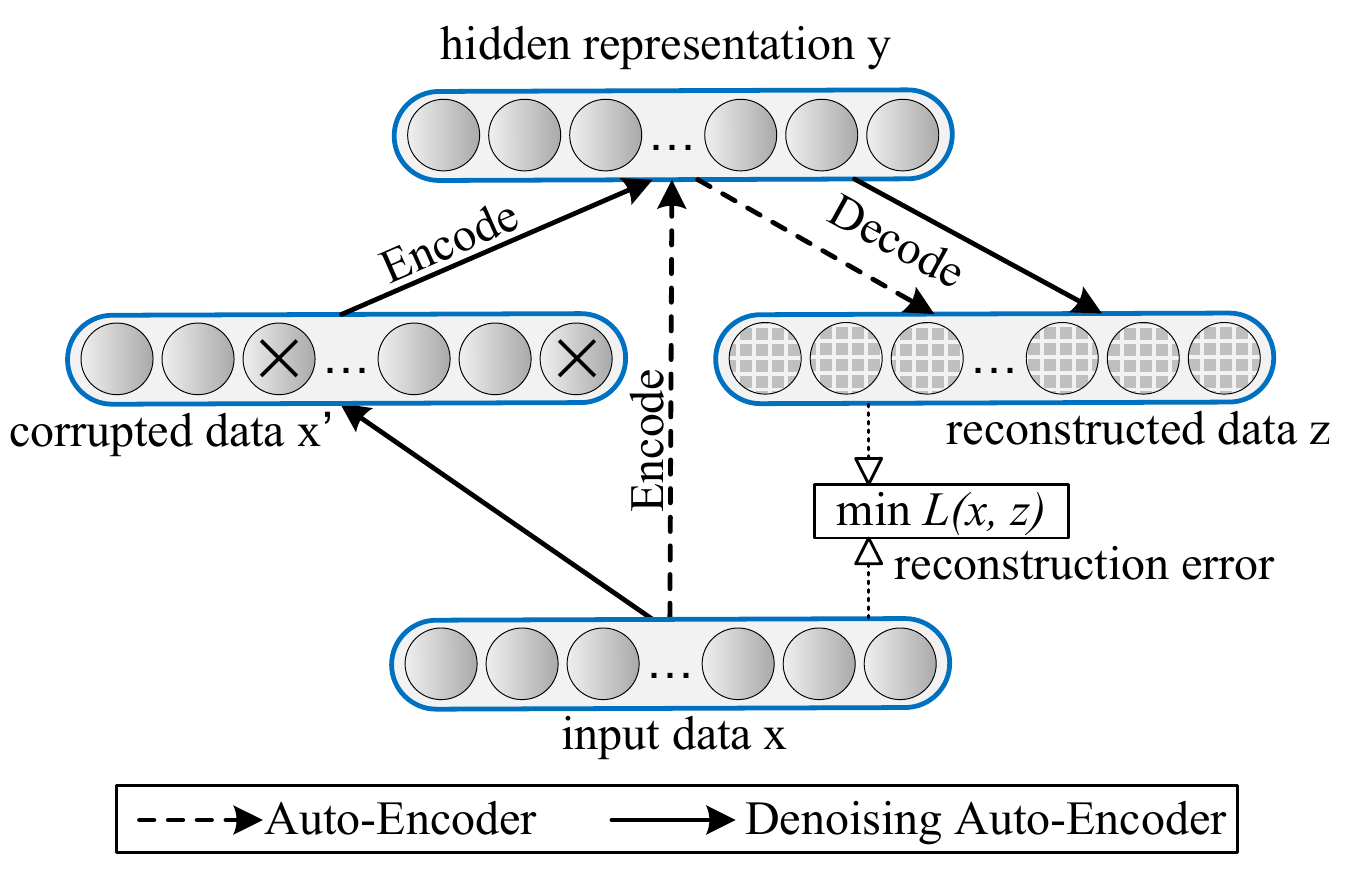}
\vspace{-0.5cm}
\caption{The Denoising Auto-Encoder (DAE) architecture. } 
\label{fig:sdae}
\end{figure}

\vspace{-0.1cm}

\subsection{SDAE-DNN pre-train strategy} 

After obtaining the superpixels and their elementary features mentioned above, the DNN classifier was trained and refined feature representation to detect building damage information. Typical deep neural network architectures include deep belief networks (DBNs) \cite{hinton2006fast}, SAEs \cite{bengio2013representation}, and stacked denoising AEs (SDAEs) \cite{vincent2010stacked}. Correspondingly, the layer-wise training models have a bunch of alternatives such as restricted Boltzmann machine (RBM) \cite{hinton2006fast}, auto-encoder (AE) \cite{bengio2013representation}, and denoising AE (DAE) \cite{vincent2010stacked}. Moreover, with respect to a supervised fine-tuned criterion, the SDAE deep architecture systematically yielded better generalization performance than DBN and SAE in several computer vision tasks. Hence, in this paper, SDAE was introduced as the corresponding deep architecture for building damage detection from VHR imagery.

\begin{figure*}[!ht]
\centering
\includegraphics[width=0.82\linewidth]{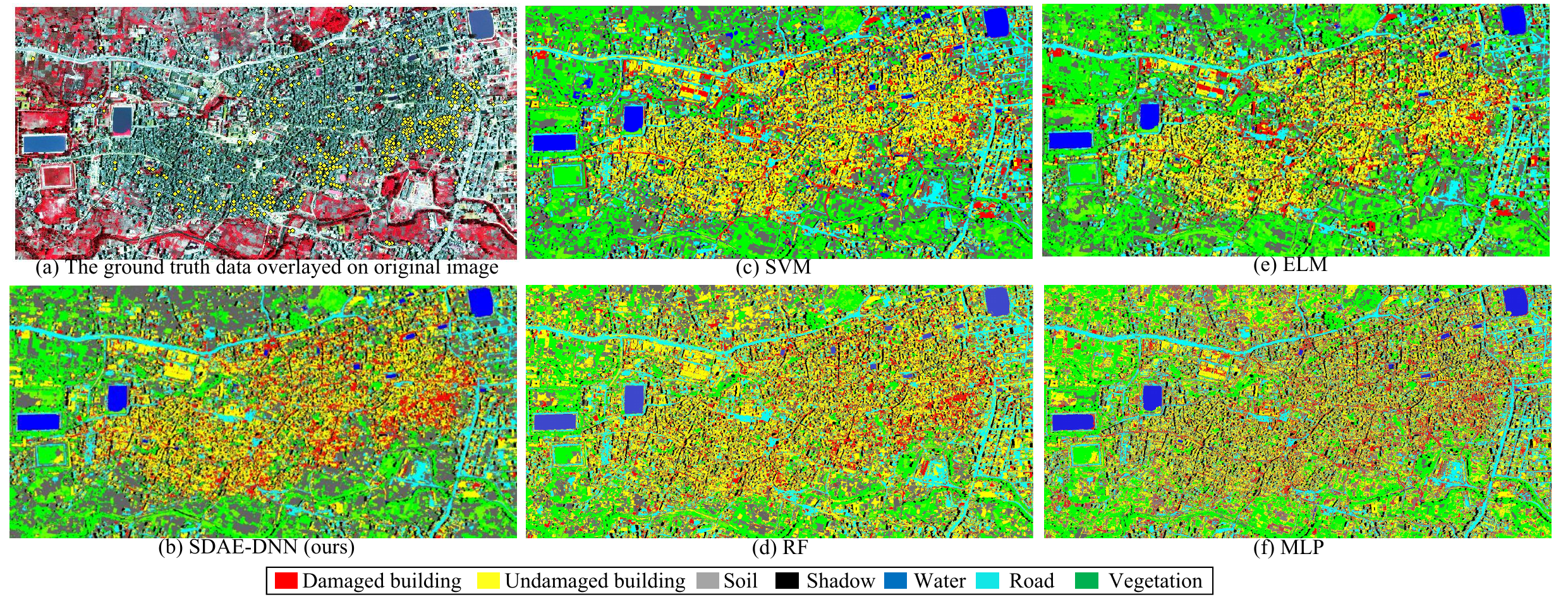}
\vspace{-0.4cm}
\caption{The ground truth data and comparison of our detection results with four alternative algorithms. (a)  The ground truth data, the yellow points indicate the damaged buildings. (b)~(f) Detection results of proposed SDAE-DNN, SVM, RF, ELM and MLP. (Red demonstrates the damaged building areas.) } 
\label{fig:results}
\end{figure*}

In the proposed method, the whole deep learning process of SDAE has 2 phases: unsupervised pre-training phase and supervised fine-tuning phase. We aim to conduct classification and exploit feature abstraction and learning simultaneously. Firstly, SDAE was constructed by stacking the modules of DAE. DAE acts as an extension of classical AE. A typical AE contains two parts: encoder and decoder. DAE reconstructs the input into a partially corrupted version, the key difference is that now $z$ is the reconstruction of $x^{’}$ rather than the original input $x$ in AE. The difference between the AE and the DAE is illustrated in Fig. \ref{fig:sdae}, where the flow of the AE is represented as dotted lines and the flow of the DAE is represented as solid lines. The encoder corresponds to a nonlinear mapping function $y = {f_\theta}(x) = s (Wx+b)$ which maps the input x from its original feature space into a hidden representation y. where W is an n*n weight matrix and b is a bias vector. $Tanh$ function is used as the active function $f(x)$. Then, the hidden representation y is mapped back to the input space through a similar decoding transformation: $z = {g_{{\theta}^{'}}}(y) = s({W^{'}}y+{b^{'}})$.
The parameters of this model are optimized by minimizing the average reconstruction error between x and z as:

\begin{equation}
\begin{aligned}
\theta^\ast,\theta^{\prime\ast}=\mathop{\arg\min}_{\theta,\theta^\prime}{\frac{1}{n}\sum\nolimits_{i=1}^{n}L\left(x_i,\ z_i\right)}
\end{aligned}
\end{equation}

\begin{equation}
\begin{aligned}
\theta^\ast,\theta^{\prime\ast}=\mathop{\arg\min}_{\theta,\theta^\prime}{\frac{1}{n}\sum\nolimits_{i=1}^{n}L\left(x_i,g_{\theta^\prime}\left(f_\theta\left(x_i\right)\right)\right)}
\end{aligned}
\end{equation}

where the loss function $L(x,z)$ is defined as reconstruction cross-entropy, i.e.,

\begin{equation}
\begin{aligned}
L\ {\left(x,\ z\right)=-\sum\nolimits_{k=1}^{n}\left\{x_klog{z_k}+\left(1-x_k\right)log\left(1-z_k\right)\right\}\ }
\end{aligned}
\end{equation}

The DAE was considered as a regularized AE. In order to discover robust features and prevent it from simply learning the identity, the AE is trained to reconstruct the input $x$ from a corrupt $x^{'}$, $x^{'} = {d_{\gamma}}(x)$, where ${d_{\gamma}}(x)$ denotes the corruption function. The original input is corrupted by randomly setting part of the inputs to be zero according to the corruption rate ${\gamma}$ (in our study ${\gamma} = 0.3$). The target function of the DAE becomes:

\begin{equation}
\begin{aligned}
\theta^\ast,\theta^{\prime\ast}=\mathop{\arg\min}_{\theta,\theta^\prime}{\frac{1}{n}\sum\nolimits_{i=1}^{n}L\left(x_i,g_{\theta^\prime}\left(f_\theta\left(d\gamma\left(x_i\right)\right)\right)\right)}
\end{aligned}
\end{equation}

The layers of SDAE are pre-trained sequentially from top to bottom. The learning target of each layer is to get the optimal weight by minimizing the target cost function defined in Equation (5). After pre-training, a logistic regression layer is added to the top layer of the SDAE. Then, the whole network is fine-tuned as an MLP using backpropagation \cite{hinton2006fast}.

\section{EXPERIMENTS AND RESULTS}

Bhaktapur, the third-largest city of Nepal, was struck on 25 April 2015 by the Magnitude 7.8 2015 Nepal earthquake. The above methodology was implemented on a 4144×2072 pixels WorldView-2 VHR imagery of Bhaktapur with 0.5 m resolution. The results are conducted with an NVIDIA GeForce GT 620 GPU. As shown in Fig. \ref{fig:results}, the ground truth data overlaid on original input images was shown in Fig. \ref{fig:results}(a), the yellow points indicate the affected buildings, which were delineated and verified manually by visual inspection, based on the damage assessment map produced by UNOSAT (United Nations Operational Satellite Applications). Then, the detection result of SDAE-DNN was demonstrated in \ref{fig:results}(b). Besides, the damage detection results by SVM, RF, ELM and MLP are listed in Fig. \ref{fig:results}(c)-\ref{fig:results}(f), respectively. 
% It should be noted that the motivation of this paper is not to investigate the performance of different machine learning algorithms, but to construct a novel superpixel-based deep learning framework for the requirements of building damage detection in VHR images. 

As for the hyperparameter setting, ﬁve-fold cross-validation (CV) is used to perform grid search and seek optimal parameters for each method. The proposed superpixel-based SDAE-DNN was implemented with 5 hidden layers, and the number of neurons in each layer is selected from (20, 50, 200, 800), and the learning rate is 0.001. Then the comparison is conducted with commonly used SVM, RF, ELM and MLP. More precisely, SVM is run with linear kernels and the ranges for optimization of penalty coefﬁcient are (0.1, 1 10, 100, 800). The parameter of RF mainly refers to the number of trees. In this study, a set of values (20, 50, 200, 800) are deﬁned for parameter selection. For ELM, the number of neurons in the hidden layer ranges from (20, 50, 200, 800). For MLP, the standard one hidden-layer MLP is implemented with hidden neurons selected from (20, 50, 200, 800).

All the accuracy evaluation of the building damage detection results is presented in Table \ref{table:1}.
The metrics used for evaluation are Kappa coefficient, precision, recall, f1-score and CV (cross-validation) accuracy.
It can be observed that our method delivers the best cross-validation accuracy of 0.877, compared with other approaches. Moreover, it yields nearly the highest precision, recall, f1-score. It is demonstrated to be effective and competitive in dealing with the issue of building damage detection based on a superpixel-based deep architecture.

\begin{table}[!h]
\centering
\caption{Accuracy comparison of various methods.}
\begin{tabular}{c|ccccc}
\hline
\multicolumn{1}{l|}{Method} & Kappa          & Precision      & Recall         & F1-score       & CV accuracy \\ \hline
SVM \cite{khoshelham2013segment}                         & 0.911          & 0.924          & \textbf{0.953} & 0.938          & 0.843                     \\
RF \cite{thomas2013automated}                          & 0.905          & 0.919          & 0.893          & 0.905          & 0.827                     \\
MLP \cite{dubois2014fast}                        & 0.813          & 0.883          & 0.794          & 0.836          & 0.776                     \\
ELM \cite{alhichri2015hierarchical}                        & 0.889          & 0.878          & 0.906          & 0.892          & 0.810                     \\
Proposed                    & \textbf{0.934} & \textbf{0.951} & 0.948          & \textbf{0.949} & \textbf{0.877}            \\ \hline
\end{tabular}
\label{table:1}
\end{table}

\vspace{-0.3cm}
Moreover, Table \ref{table:2} summarizes the computing time of mentioned methods, 
% For time efficiency comparison, the time does not include the segmentation phase, since it is the same for all the methods. 
it is observed that the SDAE-DNN causes increased training complexity, but it is worth noting that the deep architecture yields competitive and reasonable predicting time, even slightly better than SVM. Thus, experimental results demonstrated that our method not only boosted damage detection precision, but also offered a possible route to avoid the common disadvantages of time-consuming feature designing and selection in building damage detection, which outperforms conventional machine learning-based frameworks. 

\vspace{-0.3cm}

\begin{table}[!h]
\centering
\caption{COMPARISON OF THE EFFICIENCY WITH PREVIOUS WORKS.}
\resizebox{0.48\textwidth}{!}{
\begin{tabular}{c|ccccc}
\hline
Time           & SVM\cite{khoshelham2013segment}    & RF\cite{thomas2013automated}   & MLP\cite{dubois2014fast}   & ELM\cite{alhichri2015hierarchical}   & Proposed \\ \hline
Train (s)   & 13.89 & 2.72 & 16.85 & 8.90  & 195.13   \\
Inference (s) & 19.33 & 5.50 & 3.39  & 11.64 & 12.07    \\ \hline
\end{tabular}
}
\label{table:2}
\end{table}

\vspace{-0.3cm}

\section{CONCLUSIONS}

In this paper, we concern the issue of feature exploration and learning for building damage detection in VHR imagery, a superpixel segmentation based strategy integrating the SDAE-DNN architecture is proposed. Firstly, a modified region-merge segmentation method binding LBP texture is proposed, to generate meaningful superpixels better account for the spatial constraints, which could better handle the intrinsic consistency of semantics of geo-spatial objects. More importantly, our method can learn discriminate features hierarchically. The experiment results reveal that the proposed approach can achieve competitive building damage detection performance, it would play a significant role in damage assessment, and contribute to better rescue management and post-earthquake reconstruction missions.

\small
\bibliographystyle{IEEEtranN}
\bibliography{ref}

\end{document}